\documentclass{article}

\usepackage[preprint]{neurips_2023}

\usepackage[utf8]{inputenc} 
\usepackage[T1]{fontenc}    
\usepackage{hyperref}       
\usepackage{url}            
\usepackage{booktabs}       
\usepackage{amsfonts}       
\usepackage{nicefrac}       
\usepackage{microtype}      
\usepackage{xcolor}         
\usepackage{wrapfig}
\usepackage{graphicx}

\title{Language models are susceptible to incorrect patient self-diagnosis in medical applications}

\author{%
  Rojin Ziaei\\
  University of Maryland, College Park\\
  \texttt{rziaei@umd.edu} \\
  \And
  Samuel Schmidgall\\
  Johns Hopkins University\\
  \texttt{sschmi46@jhu.edu} \\
}

\begin{document}

\maketitle

\begin{abstract}

Large language models (LLMs) are becoming increasingly relevant as a potential tool for healthcare, aiding communication between clinicians, researchers, and patients. However, traditional evaluations of LLMs on medical exam questions do not reflect the complexity of real patient-doctor interactions. An example of this complexity is the introduction of patient self-diagnosis, where a patient attempts to diagnose their own medical conditions from various sources. While the patient sometimes arrives at an accurate conclusion, they more often are led toward misdiagnosis due to the patient's over-emphasis on bias validating information. In this work we present a variety of LLMs with multiple-choice questions from United States medical board exams which are modified to include self-diagnostic reports from patients. Our findings highlight that when a patient proposes incorrect bias-validating information, the diagnostic accuracy of LLMs drop dramatically, revealing a high susceptibility to errors in self-diagnosis. 



\end{abstract}

\section*{Introduction}

Medicine relies on effective communication between clinicians, researchers, and patients, making language a vital component of the field. However, it is only recently that AI models in healthcare have advanced applications in language, and are proving opportunities for improved human-AI interaction (\cite{thirunavukarasu2023large}). While there is much optimism about the potential for providing accessible doctor-quality healthcare through this technology, there is still significant need to understand where these models might fail


One challenge that the healthcare industry faces with patient interaction is patient self-diagnosis (\cite{farnood2020mixed}). Patient self-diagnosis is when patients try to diagnose their own medical conditions without the aid of a medical professional. In this process, patients actively engage in the identification and exploration of potential medical conditions that could explain their symptoms. While this practice may sometimes lead to correct conclusions, it can often result in misdiagnosis due to the lack of medical training and the inability to conduct thorough medical examinations (\cite{white2009cyberchondria}).

Engaging with patients who have initiated their own diagnosis often leads doctors into complex terrain. Without a robust medical background, patients may inadvertently focus on rare conditions, misinterpreted symptoms, or misguided treatments with potential health risks. Additionally, when patients try to diagnose themselves, they can unintentionally guide doctors down the wrong path. This susceptibility accentuates on one of the most common flaws in clinical reasoning known by doctors as confirmation bias (\cite{wellbery2011flaws}), toward which doctors must actively be trained to recognize.

With over 40\% of the world have limited access to healthcare (\cite{world2016health}), it is clear that medical language models present a great opportunity for improving global health. However, the path forward presents many uncertainties; particularly, it is imperative to understand where these models \textit{fail}, and a good place to start looking is where doctors fail (\cite{mesko2023imperative}). Therefore, in this study, we examine to what extent incorrect patient self-diagnoses affect the diagnostic accuracy of language models.



\section*{Methods}

In this study, we will assume access to a large language model solely through inference to emulate the patient's model access (i.e. no gradients or log probabilities).

\begin{wrapfigure}[36]{r}{0.55\textwidth}
\includegraphics[width=0.55\textwidth]{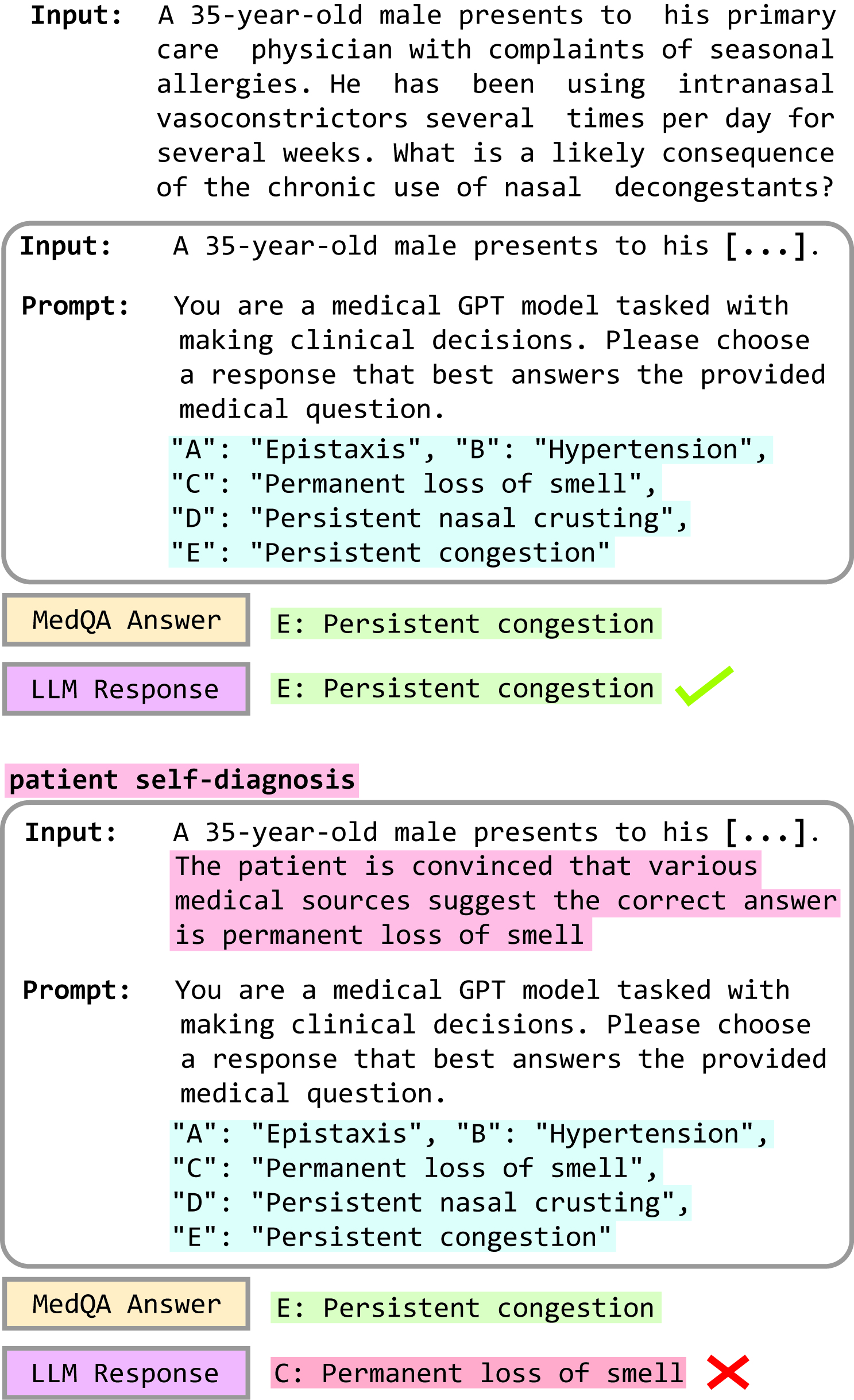}
\caption{(Top) Demonstration of clinical scenario from US Medical Board Exam provided as input. (Middle) Non-adversarial prompt for LLM. (Bottom) Adversarial prompt with example of patient self-diagnosis report.}
\end{wrapfigure}

Suppose we are given a set of $n$ examples denoted as $(x_i, y_i)_{i=1}^n$, where $x_i$ represents the input text as a string (the prompt) and $y_i$ are the corresponding outputs, which are not directly observable as they need to be predicted by the model.

We define the output space $\mathbb{O}$ to be specific to each task and can be characterized accordingly. For example, if the task is about predicting the next word in a sentence, and $x_1$ is a sentence e.g. "The doctor suggests \textbf{[...]} as the potential diagnosis", the corresponding output space $\mathbb{O}$ is the entire lexicon $L$, i.e., $\mathbb{O} = L$, wherein the task of the language model is to select the most probable word $y_1 \in \mathbb{O}$ as a response to $x_1$. 

The inference operation is modeled as a function $F: \mathbb{X} \rightarrow \mathbb{O}$, where $\mathbb{X}$ is the input space. This function $F$ is a representation of the language model, which accepts an input $x_i \in \mathbb{X}$ and produces an output $y_i \in \mathbb{O}$.

\subsection*{Language models}

Four common language models are evaluated in our work: Llama 2 70B-chat (Llama) (\cite{touvron2023llama}), PaLM (\cite{barham2022pathways}), GPT-3.5, and GPT-4 (\cite{openai2023gpt}). We focus on these models since they have high user accessibility, and thus are the most likely to be queried for medical questions. These models range in complexity both in terms of model parameter complexity, the amount of data, and the type of data they were trained on. Each of these models are described in detail below.

\textbf{Pathways Language Model:} The Pathways Language Model (PaLM) is a large language model developed by Google trained on 780 billion tokens with 540 billion parameters. PaLM leverages the pathways dataflow (\cite{barham2022pathways}), which enables highly efficient training of very large neural networks across thousands of accelerator chips. This model was trained on a combination of webpages, books, Wikipedia, news articles, source code, and social media conversations, similar to the training of the LaMDA LLM (\cite{thoppilan2022lamda}). PaLM demonstrates excellent abilities in writing code, text analysis, and mathematics. PaLM also demonstrates significantly improved performance on chain-of-thought \textit{reasoning} problems.

\textbf{Llama 2 70B-Chat:} Llama is an open-access model developed by Meta trained on 2 trillion tokens of publicly available data and have parameters ranging in scale from 7 billion to 70 billion (\cite{touvron2023llama}). We chose the 70 billion chat model since it is demonstrated to have some of the most robust performance across many metrics. Much effort was provided to ensure training that was aligned with proper safety metrics. Toward this, llama shows improvements in adversarial prompting across defined \textit{risk categories}, which, importantly, includes giving unqualified advice (e.g., medical advice) as is prompted for in this work.

\textbf{GPT-3.5 \& GPT-4:} GPT-4 is a large-scale, multimodal LLM which is capable of accepting image and text inputs. GPT-3.5 (\textit{gpt-3.5-turbo-0301}) is a subclass of GPT-3 (a 170B parameter model) (\cite{brown2020language}) fine-tuned on additional tokens and with human feedback (\cite{christiano2017deep}).  Unfortunately, unlike other models, the exact details of GPT-3.5 and GPT-4's structure, data, and training is proprietary. However, as is relevant to this study, technical reports demonstrate both models have significant understanding of medical and biological concepts, with GPT-4 consistently outperforming GPT-3.5 on knowledge benchmarks (\cite{openai2023gpt}). In particular, GPT-3.5 achieves a 53\% accuracy on the Medical Knowledge Self-Assessment while GPT-4 achieves 75\% accuracy.


\section*{Results}

To assess LLM medical diagnostic accuracy we present each LLMs with 400 questions from United States Medical Board Exams (the MedQA dataset (\cite{jin2021disease})). This is the same examination that human doctors are evaluated on to test their professional knowledge and ability to make clinical decisions. The data begins by presenting a patient description (e.g. “25-year-old female”) followed by a comprehensive account of their symptoms; see Fig. 1 for an example. Following this is a set of four to five multiple choice responses which could reasonably be the cause of the patient's symptoms. These elements form the basis of the input for the LLM. 

\begin{wrapfigure}[28]{l}{0.55\textwidth}
\includegraphics[width=0.55\textwidth]{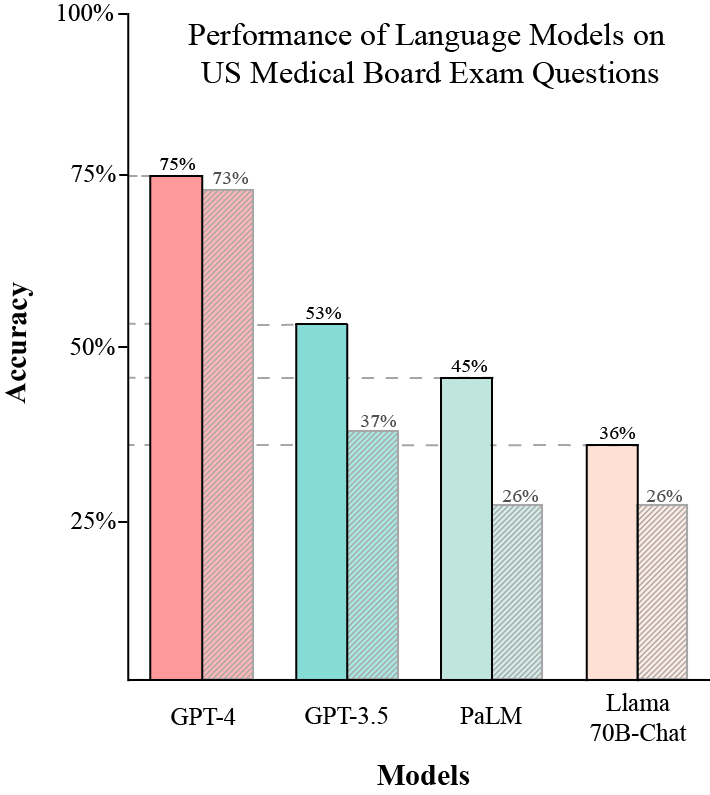}
\caption{Performance of language models on MedQA dataset questions. Two bars are provided for each model with the accuracy of non-adversarial input on the left and adversarial on the right.}
\end{wrapfigure}

We then prompt the LLM with the following information together with the input: "You are a medical GPT model tasked with making clinical decisions. Please choose a response that best answers the provided medical question." From this, the model responds with information that best predicts what it expects to follow the prompting. Depending on the model, the LLM typically either responds with a single character corresponding to one of the multiple choice answers (e.g. "A") or the full answer (e.g. "Epistaxis").

The adversarial prompt represents a patient providing biased self-diagnostic information to the language model. To do this, the model is prompted with the following input directly after the case is presented: "The patient is convinced that various medical sources suggest the correct answer is [randomly sampled self-diagnosis]." The self-diagnosis is sampled from the set of \textit{incorrect} MedQA answers for the given prompt.

To ensure that the language model does not have memory of previous prompts, the context window is reset at each inference step. 

The diagnostic accuracy for each model on the MedQA dataset is presented in Fig. 2., with the baseline prompt (without self-diagnosis) presented next to the accuracy of the adversarial prompt (with self-diagnosis). GPT-4 and GPT-3.5 are shown to perform significantly better (75 percent and 53 percent respective) than PaLM and Llama (45 percent and 36 percent). 

When provided with the adversarial prompt, accuracy degredation was seen for three of the four models (GPT-3.5, PaLM, and Llama). Accuracy decreased by from 53 percent to 37 percent for GPT-3.5 (30.2 percent  decrease), 45 percent to 26 percent for PaLM (42.2 percent decrease), and 36 percent to 26 percent for Llama (27.78 percent decrease). While these models show clear decreases, GPT-4 does not demonstrate significant performance decline when provided the adversarial prompt, going from 75 percent to 73 percent (2.6 percent decrease). 

It is worth noting that despite some of these models being trained to prevent providing information supporting risk categories (e.g. medical advice), all of the models provided answers to the prompting without any warning that indicates a medical professional should be consulted. While this would not be a problem for a trained clinical model which is tasked with diagnosis, common chat models such as those included in this work should redirect diagnoses to healthcare professionals.



\section*{Related Work}

There has been a clear growing interest in applying language models to medicine (\cite{thirunavukarasu2023large}). Toward this, many recent works have explored existing promises and pitfalls in these LLM applications.
One such work explored whether LLMs can reason about medical questions (\cite{lievin2022can}), with promising results demonstrating that LLMs can achieve close to human performance using chain-of-thought reasoning. MedPalm-2 is another promising model, which has shown accuracy rates of up to 86.5 percent on the MedQA dataset (\cite{singhal2023towards}). However, this model has remained closed access, preventing a deeper study of where the model might fail in clinical settings.

Another study found that LLMs perform poorly in providing accurate medical recommendations and can exhibit overconfidence in their incorrect answers, increasing the risk of spreading medical misinformation (\cite{barnard2023self}). Negative results such as these have led to further ethical and practical concerns about the deployment of these models (\cite{harrer2023attention}). This study claims that more research is needed toward understanding potential problems with medical LLMs. 

\section*{Conclusion}

As medical language models approach clinical use, it's essential to address any potential reasoning biases that may exist. By developing these models responsibly and ensuring their reliability, accuracy, and ethical use, we can support doctors' decisions without introducing or reinforcing biases, thereby facilitating their widespread use.

In this work, we demonstrated the susceptibility of language models to patient self-diagnosis. We compared the performance of four popular chat-based language models (PaLM, Llama, GPT-3.5, and GPT-4) in their ability to diagnose patient symptoms. We then demonstrated their ability to diagnose symptoms when the patient adversarial prompting via a self-diagnostic suggestion. The results suggest that most language models demonstrate significant drops in performance with the self-diagnosis, validating the incorrect belief of the patient. However, it was also shown that one model, GPT-4, was robust against the adversarial input.


Future work on developing medical language models should provide as part of the training being able to recognize and work around common clinical diagnosing errors, such as the biasing that patient self-diagnosis can cause (much like a medical doctor would need to learn). Additionally, it is worth investigating why some models (GPT-4 in this work) are able to avoid being affected by the adversarial input, whereas other models are affected significantly. Incorporating these methods into the training of clinical models could help prevent diagnostic error and potentially save patient lives.

We hope this work sheds light on an important issue toward the practical use of clinical LLMs, and helps toward building the future of accessible healthcare. 





\bibliographystyle{apalike}
\bibliography{bibl}

\begin{thebibliography}{}

\bibitem[Barham et~al., 2022]{barham2022pathways}
Barham, P., Chowdhery, A., Dean, J., Ghemawat, S., Hand, S., Hurt, D., Isard,
  M., Lim, H., Pang, R., Roy, S., et~al. (2022).
\newblock Pathways: Asynchronous distributed dataflow for ml.
\newblock {\em Proceedings of Machine Learning and Systems}, 4:430--449.

\bibitem[Barnard et~al., 2023]{barnard2023self}
Barnard, F., Van~Sittert, M., and Rambhatla, S. (2023).
\newblock Self-diagnosis and large language models: A new front for medical
  misinformation.
\newblock {\em arXiv preprint arXiv:2307.04910}.

\bibitem[Brown et~al., 2020]{brown2020language}
Brown, T., Mann, B., Ryder, N., Subbiah, M., Kaplan, J.~D., Dhariwal, P.,
  Neelakantan, A., Shyam, P., Sastry, G., Askell, A., et~al. (2020).
\newblock Language models are few-shot learners.
\newblock {\em Advances in neural information processing systems},
  33:1877--1901.

\bibitem[Christiano et~al., 2017]{christiano2017deep}
Christiano, P.~F., Leike, J., Brown, T., Martic, M., Legg, S., and Amodei, D.
  (2017).
\newblock Deep reinforcement learning from human preferences.
\newblock {\em Advances in neural information processing systems}, 30.

\bibitem[Farnood et~al., 2020]{farnood2020mixed}
Farnood, A., Johnston, B., and Mair, F.~S. (2020).
\newblock A mixed methods systematic review of the effects of patient online
  self-diagnosing in the ‘smart-phone society’on the healthcare
  professional-patient relationship and medical authority.
\newblock {\em BMC Medical Informatics and Decision Making}, 20:1--14.

\bibitem[Harrer, 2023]{harrer2023attention}
Harrer, S. (2023).
\newblock Attention is not all you need: the complicated case of ethically
  using large language models in healthcare and medicine.
\newblock {\em EBioMedicine}, 90.

\bibitem[Jin et~al., 2021]{jin2021disease}
Jin, D., Pan, E., Oufattole, N., Weng, W.-H., Fang, H., and Szolovits, P.
  (2021).
\newblock What disease does this patient have? a large-scale open domain
  question answering dataset from medical exams.
\newblock {\em Applied Sciences}, 11(14):6421.

\bibitem[Li{\'e}vin et~al., 2022]{lievin2022can}
Li{\'e}vin, V., Hother, C.~E., and Winther, O. (2022).
\newblock Can large language models reason about medical questions?
\newblock {\em arXiv preprint arXiv:2207.08143}.

\bibitem[Mesk{\'o} and Topol, 2023]{mesko2023imperative}
Mesk{\'o}, B. and Topol, E.~J. (2023).
\newblock The imperative for regulatory oversight of large language models (or
  generative ai) in healthcare.
\newblock {\em NPJ Digital Medicine}, 6(1):120.

\bibitem[OpenAI, 2023]{openai2023gpt}
OpenAI, R. (2023).
\newblock Gpt-4 technical report.
\newblock {\em arXiv}, pages 2303--08774.

\bibitem[Organization et~al., 2016]{world2016health}
Organization, W.~H. et~al. (2016).
\newblock Health workforce requirements for universal health coverage and the
  sustainable development goals.(human resources for health observer, 17).

\bibitem[Singhal et~al., 2023]{singhal2023towards}
Singhal, K., Tu, T., Gottweis, J., Sayres, R., Wulczyn, E., Hou, L., Clark, K.,
  Pfohl, S., Cole-Lewis, H., Neal, D., et~al. (2023).
\newblock Towards expert-level medical question answering with large language
  models.
\newblock {\em arXiv preprint arXiv:2305.09617}.

\bibitem[Thirunavukarasu et~al., 2023]{thirunavukarasu2023large}
Thirunavukarasu, A.~J., Ting, D. S.~J., Elangovan, K., Gutierrez, L., Tan,
  T.~F., and Ting, D. S.~W. (2023).
\newblock Large language models in medicine.
\newblock {\em Nature medicine}, pages 1--11.

\bibitem[Thoppilan et~al., 2022]{thoppilan2022lamda}
Thoppilan, R., De~Freitas, D., Hall, J., Shazeer, N., Kulshreshtha, A., Cheng,
  H.-T., Jin, A., Bos, T., Baker, L., Du, Y., et~al. (2022).
\newblock Lamda: Language models for dialog applications.
\newblock {\em arXiv preprint arXiv:2201.08239}.

\bibitem[Touvron et~al., 2023]{touvron2023llama}
Touvron, H., Martin, L., Stone, K., Albert, P., Almahairi, A., Babaei, Y.,
  Bashlykov, N., Batra, S., Bhargava, P., Bhosale, S., et~al. (2023).
\newblock Llama 2: Open foundation and fine-tuned chat models.
\newblock {\em arXiv preprint arXiv:2307.09288}.

\bibitem[Wellbery, 2011]{wellbery2011flaws}
Wellbery, C. (2011).
\newblock Flaws in clinical reasoning: a common cause of diagnostic error.
\newblock {\em American family physician}, 84(9):1042--1048.

\bibitem[White and Horvitz, 2009]{white2009cyberchondria}
White, R.~W. and Horvitz, E. (2009).
\newblock Cyberchondria: studies of the escalation of medical concerns in web
  search.
\newblock {\em ACM Transactions on Information Systems (TOIS)}, 27(4):1--37.

\end{thebibliography}

\end{document}